\documentclass[10pt,twocolumn,letterpaper]{article}

\usepackage[accsupp]{axessibility}  
\usepackage[pagenumbers]{cvpr} 

\usepackage{graphicx}
\usepackage{amsmath}
\usepackage{amssymb}
\usepackage{booktabs}
\usepackage{multirow}

\usepackage{amssymb}
\usepackage{pifont}
\newcommand{\xmark}{\ding{55}}%

\usepackage[pagebackref,breaklinks,colorlinks]{hyperref}

\usepackage[capitalize]{cleveref}
\crefname{section}{Sec.}{Secs.}
\Crefname{section}{Section}{Sections}
\Crefname{table}{Table}{Tables}
\crefname{table}{Tab.}{Tabs.}

\def\confName{CVPR}
\def\confYear{2023}

\def\methodName{SoY}

\def\R#1{{\mathbb{R}^{#1}}}
\def\L#1{L_{\textrm{#1}}}
\def\RR#1#2{{\mathbb{R}^{#1 \times #2}}}

\def\pose{\theta}
\def\globalrot{\gamma}
\def\npose{P}
\def\shape{\beta}

\def\betacov{{\Sigma_{\beta}}}
\def\posecov{{\Sigma_{\pose}}}
\def\posemean{\pose_{\mu}}

\def\poseT{\pose_{\textrm{T}}}
\def\globalrotT{\globalrot_{\textrm{T}}}

\def\betamean{{\mu_{\beta}}}

\def\nshape{B}

\makeatletter
\robustify\@latex@warning@no@line
\makeatother
\usepackage{authblk}
\makeatletter
\renewcommand\AB@affilsepx{, \protect\Affilfont}
\makeatother

\begin{document}

\title{Shape of You: Precise 3D shape estimations for diverse body types}
\author[1,2]{Rohan Sarkar}
\author[2]{Achal Dave}
\author[2]{Gerard Medioni}
\author[2]{Benjamin Biggs}
\affil[1]{Purdue University}
\affil[2]{Amazon}

\maketitle

\begin{abstract}
    This paper presents Shape of You (\methodName), an approach to improve the accuracy of 3D body shape estimation for vision-based clothing recommendation systems. While existing methods have successfully estimated 3D poses, there remains a lack of work in precise \emph{shape} estimation, particularly for diverse human bodies. To address this gap, we propose two loss functions that can be readily integrated into parametric 3D human reconstruction pipelines. Additionally, we propose a test-time optimization routine that further improves quality. Our method improves over the recent SHAPY~\cite{SHAPY2022} method by 17.7\% on the challenging SSP-3D dataset~\cite{STRAPS2018BMVC}. We consider our work to be a step towards a more accurate 3D shape estimation system that works reliably on diverse body types and holds promise for practical applications in the fashion industry.
\end{abstract}

\section{Introduction}
\label{sec:intro}

Clothing retailers have been designing interactive experiences that make clothing recommendations based on customer selfie photos, which contain valuable visual cues pertaining to body shape. Approaches promise to transform digital shopping experiences by empowering customers to shop more confidently without physical access to the inventory and by reducing the rate of clothing returns. However, estimating shape characteristics from 2D images of diverse human bodies remains a challenging problem. 

Techniques for image-based 3D human reconstruction present compelling opportunities in this space. These approaches disentangle the space of human bodies into \emph{shape} deformations that control relative body proportions, and \emph{pose} deformations that control the position and orientation of limbs. Precisely estimating a customer's \emph{shape} enables valuable fashion applications, such as being able to predict suitable clothing sizes or identify flattering garments. Pose deformations are typically nuisance factors, but they must still be modeled carefully to avoid noisy shape estimates.

Despite significant recent progress for image-based 3D human reconstruction, most improvements have been made in robust 3D pose estimation. Precisely estimating \emph{shape} characteristics, particularly for plus-size body types, remains a challenging problem, in part due to the scarcity of training data. \Cref{fig:SPINDP} compares our proposed approach to that of Choutas et al.~\cite{SHAPY2022}. We achieve improved shape estimates despite training without annotated body measurements or semantic textual attributes that are used by their method. Next, we summarize our paper's key contributions:

\begin{figure}[t]
    \centering
    \includegraphics[width=0.5\textwidth]{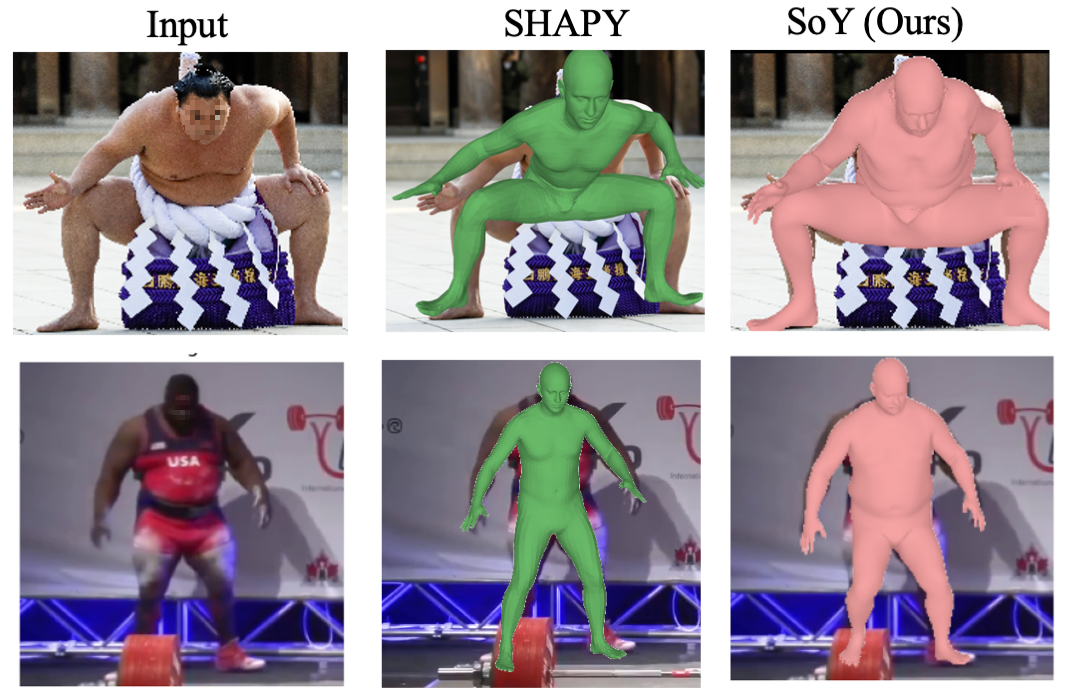}
    \caption{Our \methodName{} method generates improved body shape estimates compared to prior work~\cite{SHAPY2022}.}
    \label{fig:SPINDP}
    \vspace{-0.1in}
\end{figure}

\begin{enumerate}
    \item We propose \methodName{}, a novel 3D human reconstruction method that incorporates specific loss functions to promote detailed 3D shape recovery, particularly for diverse humans shapes. Our method includes (a) a 2D loss based on dense correspondences~\cite{DENSEPOSE} between the 3D mesh and foreground person -- an improved shape-specific signal over 2D keypoints~\cite{SPIN2019,SMPLify16} or 2D silhouettes~\cite{SMAL} -- and (b) a 3D loss that promotes shape-specific vertex alignment by subtracting pose effects.
    \item We propose a refinement step which further improves the quality of shape recovery. This procedure is particularly useful when testing on humans with body shapes that are under-represented in the training distribution.
    \item We show that \methodName{} with refinement achieves an improvement of 17.7\%  over Choutas et al.~\cite{SHAPY2022} on SSP-3D without \emph{any body measurements or semantic attributes for training}. Additionally, we show our method performs on-par when the refinement step is omitted.
\end{enumerate}

\section{Related Work}

The problem of recovering 3D shape and pose for articulated subjects has received significant attention in the literature. 
One common approach is to fit 3D morphable body models (3DMM) to the subjects of interest. Early methods relied on iterative optimization algorithms~\cite{SMPLify16} to align 3DMMs with 2D observations, while recent works~\cite{HMR18, GRAPHCMR} estimate 3DMM parameters with direct learning-based regression. Our work builds on the hybrid approaches~\cite{SPIN2019} which integrate an iterative optimization loop for training the deep network that performs the regression. All of these methods fall short in accurately estimating shape for plus-size individuals. In this paper, we propose loss functions and a refinement procedure to overcome this limitation. 

Another emerging trend is to reconstruct 3D articulated subjects without an explicit 3D template prior~\cite{PIFU, PIFUHD, PHORHUM, SNARF}. While precise shapes can be estimated, they rely heavily on paired 3D training data which is limited for diverse body shapes and tend to entangle clothing in the reconstructed output. Furthermore, it is unclear how best to integrate these neural network-encoded shape representations into downstream fashion applications.

Recent animal reconstruction literature has made progress for complex shape categories~\cite{SMAL, WLDO, CGAS, BARC:2022}. Our paper takes particular inspiration from BANMo~\cite{Banmo2022} who fit to dense correspondences tracked within a video sequence.

However, the most relevant are techniques that reconstruct diverse human shapes. Some methods~\cite{STRAPS2018BMVC, PROBPOSE} augment existing 3D training datasets with synthetically-generated diverse body shapes. However, this is achieved using in-the-wild datasets~\cite{3dpw,up3d} of 3D joint angles which is more than we require here. SHAPY~\cite{SHAPY2022} proposes a method to recover 3D shapes for diverse fashion models. However, they train using annotated measurements and semantic textual attributes -- a requirement we overcome in this work.

\section{Preliminaries}

Before introducing our method, we cover the necessary background starting with SMPL.

\textbf{SMPL: }
SMPL is a 3D model of the human body parameterized by ($23 \times 3)$ axis-angle rotations $\pose \in \R{\npose}$ of the limbs, shape coefficients $\shape \in \R{\nshape}$ that control body proportions, and a global rotation parameter $\globalrot \in \R{3}$. SMPL is supplied with a linear blend skinning function $S: (\pose, \shape, \globalrot) \mapsto V$, which generates a set of vertex positions $V \in \RR{6890}{3}$ of a 3D mesh.

\textbf{Predicting SMPL parameters from a single monocular image: }
For a person in input image $I$, the 3D reconstruction task is to estimate their SMPL parameters $(\pose, \shape, \globalrot)$. Modern algorithms \cite{HMR18, SPIN2019} train a deep network $G(I)$ that directly estimates these parameters as well as the translation $t \in \R{3}$ of the perspective camera with fixed parameters. Methods are trained on a mix of datasets with various levels of annotation. Vertex losses can be applied when dense 3D scans are available. Otherwise, a fixed linear regressor $J: V \mapsto X$ translates SMPL mesh vertices to 3D joints enabling comparison with sparse 3D annotations. Furthermore, 2D joint losses can be formed using a function $\pi_{t}(X)$ that projects 3D joints to the image plane.

\section{Proposed method}

\begin{figure}[t]
    \centering
    \includegraphics[width=0.5\textwidth]{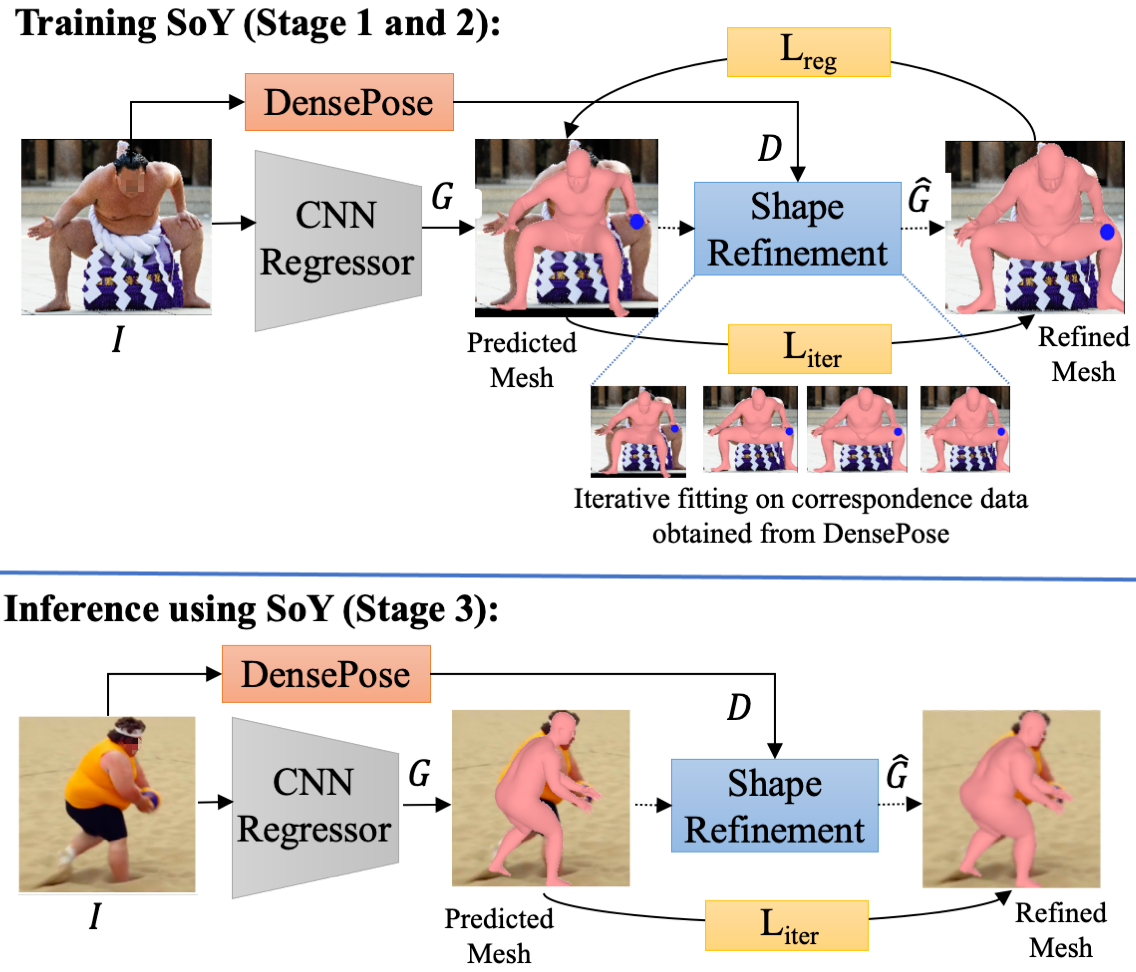}
    \caption{System Overview. 
    {\bf Stage 1}: A CNN Regressor is trained to estimate the SMPL parameters $G$ using the $L_{reg}$ loss (ref. \cref{eq:reg}).
    {\bf Stage 2}: The Shape Refinement module is initialized with $G$ and runs an iterative optimization process with loss $\L{iter}$  (ref. \cref{eq:iter}) based on dense 3D correspondences $D$ to produce updated parameters $\hat{G}$, used as pseudo-ground truth in the next epoch.
    {\bf Stage 3}: At inference time, the trained regressor's prediction $G$ is refined using $L_{iter}$ to produce the final fit.}
    \label{fig:system_overview}
    \vspace{-0.1in}
\end{figure}

We begin with a function that implements $G(I)$ as described above. As shown in \cref{fig:system_overview}, we start with the training protocol of SPIN~\cite{SPIN2019} which contains the following steps:

\begin{list}{$\bullet$}{\leftmargin=1em \itemindent=0em}
    \item \textbf{Stage 1}. Train a feed-forward neural network~\cite{HMR18} $G(I)$ to regress SMPL parameters $G$ using a loss $\L{reg}$.
    \item \textbf{Stage 2}. During training, refine the network's predictions $G$ with an energy-minimization framework (based on ~\cite{SMPLify16}) with loss $\L{iter}$. Use the refined predictions $\hat{G}$ as pseudo ground-truth for the feed-forward network in the next epoch.
\end{list}

However, as shown later in \Cref{tab:baselines}, SPIN produces poor shape estimates for plus-sized bodies. To ameliorate this, our \methodName{} method extends SPIN  introducing two additional loss functions $\L{dp}$ and $\L{tpose}$ which we add to $\L{reg}$ and $\L{iter}$. Further, we propose an additional test-time stage:

\begin{list}{$\bullet$}{\leftmargin=1em \itemindent=0em}
    \item \textbf{Stage 3}. At test-time, run another pass of energy minimization with loss $\L{iter}$ to refine the test-time predictions. 
\end{list}

Note that the energy minimization \emph{does not use any ground-truth data}; the refinement process is based on correspondence data that is predicted from the image. The next section formally introduces our proposed losses.

\subsection{Shape-specific losses}
\label{sec:losses}
In this section, we propose two loss functions that promote strong shape alignment during training.

\begin{figure}[t]
    \centering
    \includegraphics[width=0.5\textwidth]{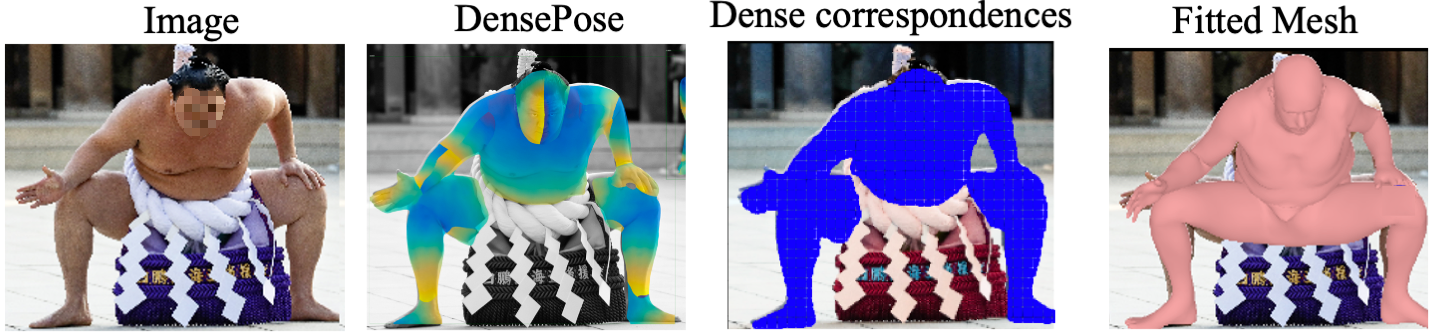}
    \caption{
    This figure shows the dense image-to-mesh correspondences derived from DensePose. The mesh fitted to these dense correspondences using $L_{iter}$ (ref. \cref{eq:iter}) is shown on the right.}
    \label{fig:dense_corr}
\end{figure}

\noindent 
\textbf{Dense correspondences: } Our first loss function promotes alignment between a set of per-pixel mesh-to-image correspondences. As shown in \cref{fig:dense_corr}, we use the popular DensePose~\cite{DENSEPOSE} method to generate a map $D_{G} \in \RR{HW}{3}$ which encodes a 3D point on the posed mesh surface for every pixel $x$ in an image $I$ of size $H \times W$. 
\begin{equation}\label{eqt:densepose}
    \L{dp}(G; D) = \sum^{H \times W}_{x} || \pi_{t}(D_{G}(x)) - x ||^{2}
\end{equation}

\noindent 
\textbf{Pose-invariant vertex alignment: }
Our second loss function promotes alignment between predicted and refined mesh vertices.
Note that we experimented with increasing the weight for SPIN's existing loss between shape coefficients $\beta, \hat{\beta}$ but found this leads to poor results. Instead, we design a loss that penalizes variations between mesh vertices $v$ when \emph{generated in T-Pose} -- that is, with any pose effects removed. Here, $\poseT=\globalrotT=0$ are the pose and global rotation parameters for the model in T-Pose.
\begin{equation}\label{eqt:tpose}
    \L{tpose}(G; \hat{G}) = \sum_{v \in V} || S_{v}(\poseT, \shape, \globalrotT) - S_{v}(\poseT, \hat{\shape}, \globalrotT) ||^{2}
\end{equation}
\noindent 
\textbf{Total losses: }
The feed-forward regression process is therefore supervised by the following losses where each has a constant weight. $\hat{G}$ are refined SMPL parameters and $\hat{Y}$ are 2D joints predicted with OpenPose~\cite{OPENPOSE}:

\begin{equation}\label{eq:reg}
\begin{aligned}
    \L{reg}(G;& \hat{G},\hat{Y}) = \L{mesh}(G; \hat{G}) + \L{3D}(G; \hat{G}) \\
    &+ \L{2D}(\pi_{t}(X); \hat{Y}) 
     + \L{tpose}(G; \hat{G})
\end{aligned}
\end{equation}

The iterative optimization process is supervised by the following losses, where $\betamean, \betacov, \posemean, \posecov$ are mean vectors and covariance matrices for shape and pose parameters respectively. These are provided as part of the SMPL model.
\vspace{-1em}
\begin{equation}\label{eq:iter}\begin{aligned}
    \L{iter}(G; D, \mu, \Sigma) &= \L{dp}(G; D) + \L{prior}(G; \mu, \Sigma)
\end{aligned}
\end{equation}

The losses $\L{mesh}, \L{3D}, \L{2D}$ and $\L{prior}$ were introduced in SPIN and we refer the reader to \cite{SPIN2019} for details.

\subsection{Implementation details}
\label{sec:impl}

    We initialize the feed-forward regressor with weights provided by SPIN~\cite{SPIN2019} and set the loss weights as follows.
    \textbf{Stage 1}: $\lambda_{mesh}=0.1, \lambda_{3D}=1.0, \lambda_{2D}=1.0, \lambda_{tpose}=0.1$. We use the ADAM optimizer with LR $=5e^{-5}$.\\ \textbf{Stage 2}: We fix $\globalrot, t$ from Stage 1, and optimize for $\shape, \pose$ with $L_{iter}$ for 250 iterations per epoch. We set $\lambda_{dp}=99.9, \lambda_{prior,\pose}=1.0$ and $\lambda_{prior,\shape}=5.0$.\\ \textbf{Stage 3}. For the SSP3D dataset \cite{STRAPS2018BMVC}, the optimal weights for refinement are found to be $\lambda_{prior,\pose}=\lambda_{prior,\shape}=25.0$. Refinement takes approximately 8s per image.
    
\begin{figure*}[h] 
    \centering
    \begin{minipage}{0.428\textwidth}
    \fbox{\includegraphics[width=\textwidth]{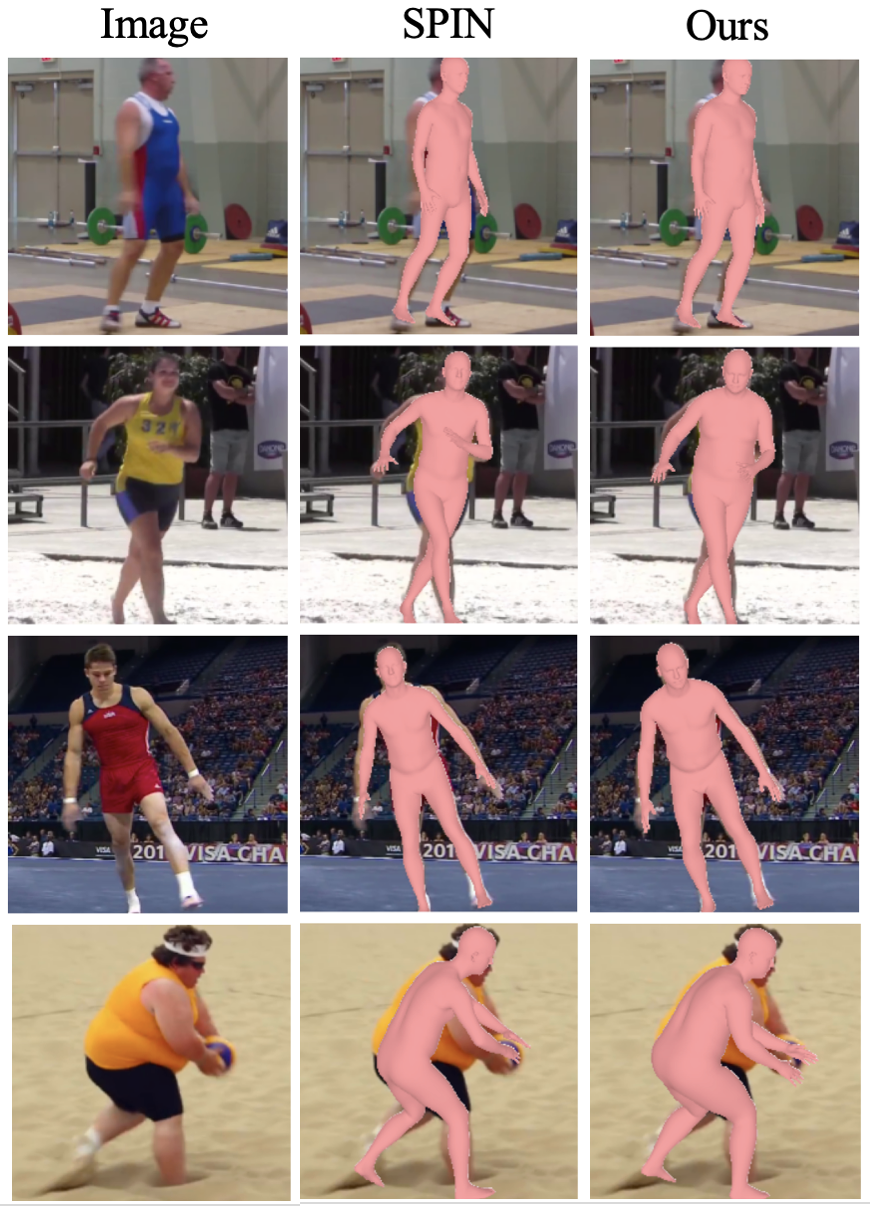}}
    
\end{minipage}
\hfill
    \begin{minipage}{0.55\textwidth}
    \centering
    \fbox{\includegraphics[width=0.98\textwidth]{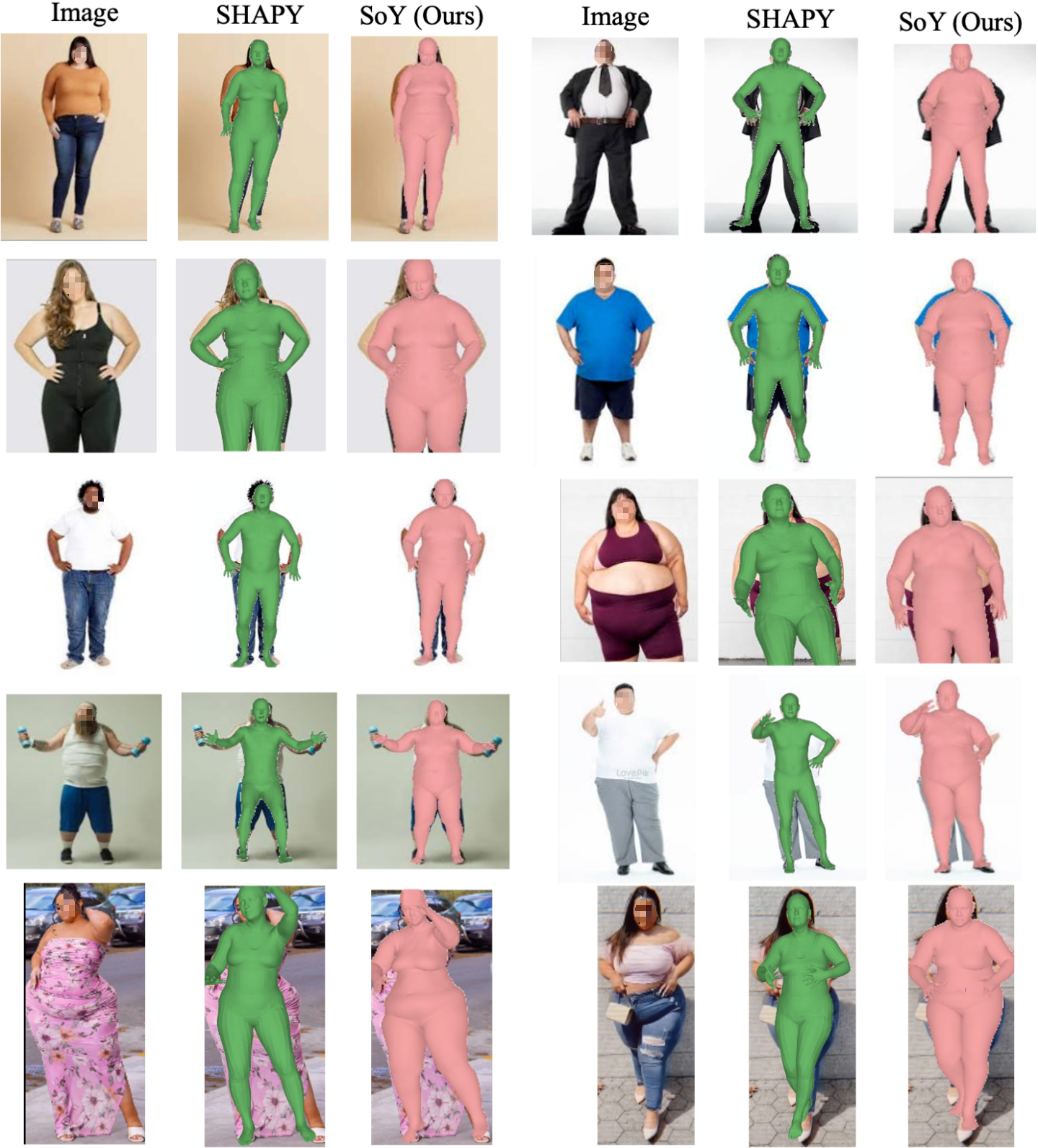}}
    
    \end{minipage}
    \vspace{-0.07in}
    \caption{Comparison of results generated using SHAPY and \methodName{} on (left) diverse body types of sports athletes from SSP3D and (right) high BMI body types in a variety of different clothing.}
    \label{fig:results}   
\end{figure*}

\section{Experiments}

This section describes a quantitative and qualitative evaluation against competitive baselines.  

\subsection{Baselines, datasets and evaluation protocol}

We evaluate our approach against three state-of-the-art techniques: HMR~\cite{HMR18}, SPIN~\cite{SPIN2019} and SHAPY~\cite{SHAPY2022}. We do not compare to methods~\cite{STRAPS2018BMVC, PROBPOSE} that require in-the-wild datasets~\cite{3dpw,up3d} of 3D joint angles for training. 

We pretrain on the SPIN~\cite{SPIN2019} training datasets and fine-tune on the Model Agency dataset~\cite{SHAPY2022} with generated OpenPose~\cite{OPENPOSE} joints and DensePose maps~\cite{DENSEPOSE}. We do not use body measurement data or linguistic body shape annotations.

Our quantitative evaluation is based on the SSP-3D dataset~\cite{STRAPS2018BMVC} which contains 311 images of sportspeople with diverse shapes in tight-fitting clothes. 
Following SHAPY~\cite{SHAPY2022}, we report per-vertex error in millimeters in T-pose after scale correction (PVE-T-SC) and mean intersection-over-union (mIoU) between predicted and ground-truth 2D silhouettes. We also provide qualitative results using a set of diverse bodies we sourced online.




\begin{table}[t]
\centering
\begin{tabular}{@{}lcc@{}}
\toprule
Method & PVE-T-SC ($\downarrow$) & mIoU ($\uparrow$) \\ \midrule
HMR  & 22.9 & 0.69 \\
SPIN  & 22.2 & 0.70 \\
SHAPY  & 19.2 & 0.71 \\
\hline
\methodName{} (Ours) & {\bf 15.8} & \textbf{0.76} \\
\bottomrule
\end{tabular}
\caption{Comparison of scaled mean vertex-to-vertex error in T-pose (PVE-T-SC) and mean intersection-over-union (mIoU) for our \methodName{} against competitive baselines.}
\label{tab:baselines}

\vspace{1em}

\centering
\begin{tabular}{@{}cccc@{}}
\toprule
$\L{tpose}$ & Stage 3 & PVE-T-SC ($\downarrow$) & mIoU ($\uparrow$) \\ \midrule
\checkmark & \checkmark & {\bf 15.8} & \textbf{0.76} \\
\checkmark & \xmark & 19.1 & 0.65 \\ 
\xmark & \xmark & 19.6 & 0.64\\ 
\bottomrule
\end{tabular}
\caption{Ablation Study on SSP-3D test set. We compare our full method (R1) with a version of our method trained without the Stage 3 refinement step (R2) and without the $\L{tpose}$ loss (R3).}
\label{tab:abl_tpose}
\vspace{-0.1in}
\end{table}

\subsection{Results}

\noindent 
\textbf{Baselines.} \Cref{tab:baselines} compares \methodName{} 
against the baselines. Our approach outperforms all three baselines and does not require the semantic linguistic labels or body measurement labels employed by SHAPY. 
 
\noindent 
\textbf{Qualitative.} \cref{fig:results} shows examples of our method running on a set of images of diverse body shapes from SSP3D, and on plus-size individuals in different clothing we downloaded online. It can be seen that our model improves the accuracy of shape estimation on a diverse set of bodies as compared to prior work \cite{SHAPY2022}. The improvements are most noticeable for plus-sized individuals whose body characteristics are most under-represented in the training datasets. As shown in the bottom row, our refinement process also improves the quality of the estimated pose.

\noindent 
\textbf{Ablation Study.} \Cref{tab:abl_tpose} demonstrates that the Stage 3 refinement step leads to a 17.3\% improvement on PVE-T-SC and our proposed T-pose loss improves the performance of our feed-forward network by 2.6\%. Note that our method performs on par with SHAPY with the test-time refinement step removed.

\section{Conclusion}
This paper explores an important and understudied problem of estimating the 3D shape of humans whose body characteristics are under-represented in computer vision datasets. We demonstrate two shape-specific loss functions and a test-time iterative refinement technique that improves the quality of shape estimates for this group, as tested on the challenging SSP-3D dataset. We achieve this without using semantic text attributes or body measurements for training.

{\small
\bibliographystyle{ieee_fullname}
\bibliography{egbib}

\begin{thebibliography}{10}\itemsep=-1pt

\bibitem{PHORHUM}
Thiemo Alldieck, Mihai Zanfir, and Cristian Sminchisescu.
\newblock Photorealistic monocular 3d reconstruction of humans wearing
  clothing.
\newblock In {\em Proc. CVPR}, 2022.

\bibitem{WLDO}
Benjamin Biggs, Oliver Boyne, James Charles, Andrew Fitzgibbon, and Roberto
  Cipolla.
\newblock {W}ho left the dogs out?: {3D} animal reconstruction with expectation
  maximization in the loop.
\newblock In {\em Proc. ECCV}, 2020.

\bibitem{CGAS}
Benjamin Biggs, Thomas Roddick, Andrew Fitzgibbon, and Roberto Cipolla.
\newblock {C}reatures great and {SMAL}: {R}ecovering the shape and motion of
  animals from video.
\newblock In {\em Proc. ACCV}, 2018.

\bibitem{SMPLify16}
Federica Bogo, Angjoo Kanazawa, Christoph Lassner, Peter Gehler, Javier Romero,
  and Michael Black.
\newblock Keep it smpl: Automatic estimation of 3d human pose and shape from a
  single image.
\newblock In {\em Proc. ECCV}, pages 561--578, 10 2016.

\bibitem{OPENPOSE}
Z. {Cao}, G. {Hidalgo Martinez}, T. {Simon}, S. {Wei}, and Y.~A. {Sheikh}.
\newblock Openpose: Realtime multi-person 2d pose estimation using part
  affinity fields.
\newblock In {\em TPAMI}, 2019.

\bibitem{SNARF}
Xu Chen, Yufeng Zheng, Michael~J Black, Otmar Hilliges, and Andreas Geiger.
\newblock Snarf: Differentiable forward skinning for animating non-rigid neural
  implicit shapes.
\newblock In {\em Proc. ICCV}, 2021.

\bibitem{SHAPY2022}
Vasileios Choutas, Lea Muller, Chun-Hao~P. Huang, Siyu Tang, Dimitris Tzionas,
  and Michael~J. Black.
\newblock Accurate 3d body shape regression using metric and semantic
  attribute.
\newblock In {\em Proc. CVPR}, June 2022.

\bibitem{DENSEPOSE}
R{\i}za~Alp G{\"u}ler, Natalia Neverova, and Iasonas Kokkinos.
\newblock Densepose: Dense human pose estimation in the wild.
\newblock In {\em Proc. CVPR}, pages 7297--7306, 2018.

\bibitem{HMR18}
Angjoo Kanazawa, Michael~J. Black, David~W. Jacobs, and Jitendra Malik.
\newblock End-to-end recovery of human shape and pose.
\newblock In {\em Proc. CVPR}, 2018.

\bibitem{SPIN2019}
Nikos Kolotouros, Georgios Pavlakos, Michael~J Black, and Kostas Daniilidis.
\newblock Learning to reconstruct 3d human pose and shape via model-fitting in
  the loop.
\newblock In {\em Proc. ICCV}, 2019.

\bibitem{GRAPHCMR}
Nikos Kolotouros, Georgios Pavlakos, and Kostas Daniilidis.
\newblock Convolutional mesh regression for single-image human shape
  reconstruction.
\newblock In {\em Proc. CVPR}, 2019.

\bibitem{up3d}
Christoph Lassner, Javier Romero, Martin Kiefel, Federica Bogo, Michael~J.
  Black, and Peter~V. Gehler.
\newblock Unite the people: Closing the loop between 3d and 2d human
  representations.
\newblock In {\em Proc. CVPR}, July 2017.

\bibitem{BARC:2022}
Nadine Rueegg, Silvia Zuffi, Konrad Schindler, and Michael~J. Black.
\newblock Barc: Learning to regress 3d dog shape from images by exploiting
  breed information.
\newblock In {\em Proc. CVPR}, 2022.

\bibitem{PIFU}
Shunsuke Saito, Zeng Huang, Ryota Natsume, Shigeo Morishima, Angjoo Kanazawa,
  and Hao Li.
\newblock Pifu: Pixel-aligned implicit function for high-resolution clothed
  human digitization.
\newblock In {\em Proc. ICCV}, October 2019.

\bibitem{PIFUHD}
Shunsuke Saito, Tomas Simon, Jason Saragih, and Hanbyul Joo.
\newblock Pifuhd: Multi-level pixel-aligned implicit function for
  high-resolution 3d human digitization.
\newblock In {\em Proc. CVPR}, 2020.

\bibitem{STRAPS2018BMVC}
Akash Sengupta, Ignas Budvytis, and Roberto Cipolla.
\newblock Synthetic training for accurate 3d human pose and shape estimation in
  the wild.
\newblock In {\em Proc. BMVC}, September 2020.

\bibitem{PROBPOSE}
Akash Sengupta, Ignas Budvytis, and Roberto Cipolla.
\newblock {Probabilistic 3D Human Shape and Pose Estimation from Multiple
  Unconstrained Images in the Wild}.
\newblock In {\em Proc. CVPR}, June 2021.

\bibitem{3dpw}
Timo von Marcard, Roberto Henschel, Michael~J. Black, Bodo Rosenhahn, and
  Gerard Pons-Moll.
\newblock Recovering accurate 3d human pose in the wild using imus and a moving
  camera.
\newblock In {\em Proc. ECCV}, September 2018.

\bibitem{Banmo2022}
Gengshan Yang, Minh Vo, Natalia Neverova, Deva Ramanan, Andrea Vedaldi, and
  Hanbyul Joo.
\newblock Banmo: Building animatable 3d neural models from many casual videos.
\newblock In {\em Proc. CVPR}, pages 2863--2873, June 2022.

\bibitem{SMAL}
Silvia Zuffi, Angjoo Kanazawa, David Jacobs, and Michael~J. Black.
\newblock {3D} menagerie: Modeling the {3D} shape and pose of animals.
\newblock In {\em Proc. CVPR}, July 2017.

\end{thebibliography}
}

\end{document}